%% file: ERRL.tex
\documentclass[english]{DDCLSconf}
\usepackage[comma,numbers,square,sort&compress]{natbib}
\usepackage{epstopdf}
\usepackage{amsfonts}
\usepackage{amsmath}
\usepackage{graphicx}
\usepackage{algorithm}
\usepackage{algpseudocode}
\usepackage{booktabs}
\usepackage{hyperref}
\usepackage{amssymb}

\begin{document}
\pagestyle{plain}

\title{ELO-Rated Sequence Rewards: Advancing Reinforcement Learning Models}

\author{Qi Ju\aref{hust},
        Falin Hei\aref{hust},
        Zhemei Fang\aref{hust},
        Yunfeng Luo\aref{hust}}



\affiliation[hust]{School of Artificial Intelligence and Automation,
        Huazhong University of Science and Technology, Wuhan 430074, P.~R.~China
        \email{juqi@hust.edu.cn}~~~~~{heifalin@hust.edu.cn}}

\maketitle

\begin{abstract}
Reinforcement Learning (RL) heavily relies on the careful design of the reward function. However, accurately assigning rewards to each state-action pair in Long-Term Reinforcement Learning (LTRL) tasks remains a significant challenge. As a result, RL agents are often trained under expert guidance. Inspired by the ordinal utility theory in economics, we propose a novel reward estimation algorithm: ELO-Rating based Reinforcement Learning (ERRL). This approach features two key contributions. First, it uses expert preferences over trajectories rather than cardinal rewards (utilities) to compute the ELO rating of each trajectory as its reward. Second, a new reward redistribution algorithm is introduced to alleviate training instability in the absence of a fixed anchor reward. In long-term scenarios (up to 5000 steps), where traditional RL algorithms struggle, our method outperforms several state-of-the-art baselines. Additionally, we conduct a comprehensive analysis of how expert preferences influence the results.
\end{abstract}



    \section{Introduction}\label{sec:introduction}
    \input{p1introduction}

    \section{Preliminaries}\label{sec:preliminaries}
    \input{p2preliminaries}

    \section{Methodology}\label{sec:methodology}
    \input{p4methodology}

    \section{Experiments}\label{sec:experiments}
    \input{p5experiments}

    \section{Related Work}\label{sec:related-work}
    \input{p3related_work}

    \section{Conclusion and Future Work}\label{sec:summary-and-conclusion}
    \input{p7summary_and_conclusion}

    \clearpage

\input{p8references}
    \clearpage

    \section*{Appendix A}\label{sec:appendix-a}
    \input{p91Appendix_A}

    \section*{Appendix B}\label{sec:appendix-b}
    \input{p92Appendix_B}

    \section*{Appendix C}\label{sec:appendix-c}
    \input{p93Appendix_C}

\end{document}

%% file: p1introduction.tex
Deep Reinforcement Learning (RL) has witnessed significant advancements across diverse domains, such as Atari games~\cite{ref7}, Go~\cite{ref2,ref5}, poker~\cite{ref12}, and video gaming~\cite{ref4,ref6}. However, real-world applications often encounter sparse environmental feedback. For instance, in a game of Go, which may involve around \(N \approx 200\) decision states, the reward signal is solely available at the final state. This sparsity necessitates meticulous design of reward functions for RL algorithms. Moreover, the expansive state spaces in many games impede agents from effectively learning through self-exploration.

In response, it is common practice to either engage domain experts in reward function design~\cite{ref29} or employ imitation learning for pre-training to guide agents in learning optimal policies~\cite{ref1,ref30}. Nevertheless, reward functions crafted by human experts are confronted with three primary challenges: First, potential errors or latent flaws within the reward function can be exploited by RL agents to obtain spurious rewards; Second, expert-designed functions may impose an upper bound on the performance of RL agents, hindering them from surpassing human expertise; Third, tackling large-scale problems usually demands a team of experts, which not only escalates costs but also often results in policies lacking uniqueness due to the requirement for expert consensus.

To address these challenges, we draw inspiration from the ordinal utility theory in economics, which posits that \textit{decision-makers do not compute the exact numerical value of each choice, but rather derive decisions by comparing the relative merits of strategies}~\cite{ref14}. This motivates the hypothesis that RL agents may not require precise cardinal rewards for effective learning—instead, preferences over trajectories could suffice to train agents with comparable or superior performance.

\begin{figure*}
    \centering
    \includegraphics[width=\textwidth]{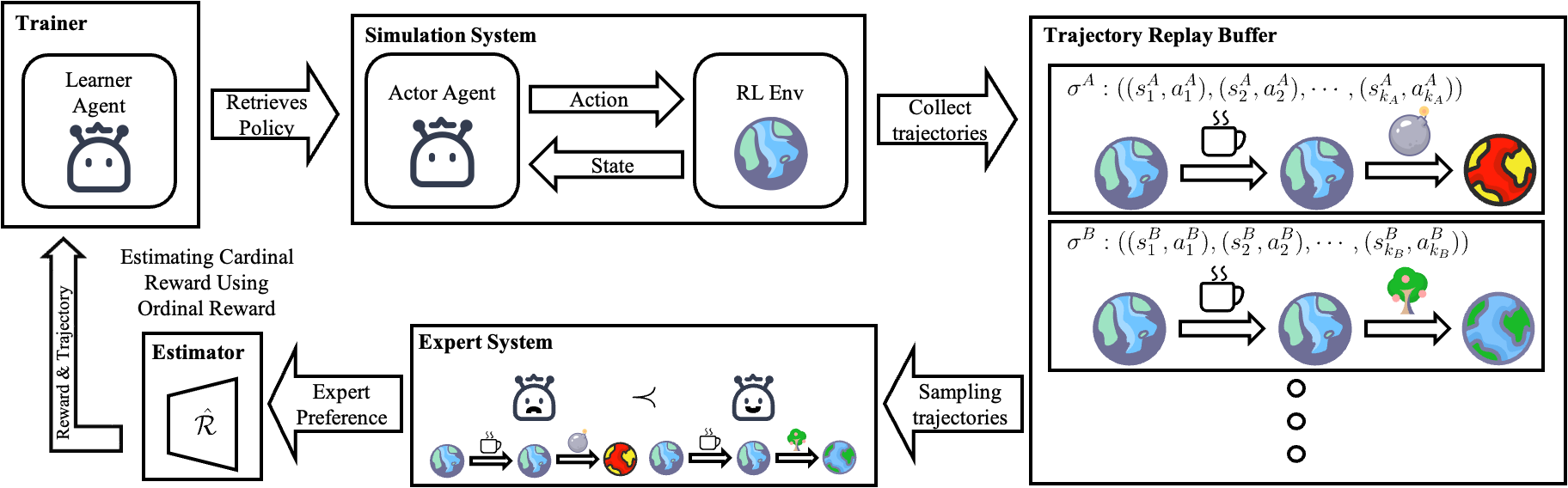}
    \caption{
        Unlike classical RL, preference-based RL has no direct reward.
        Expert  judgment is required  when sampling trajectories from the replay buffer,
        and reward are obtained through expert judgment.
    }\label{fig:fig1}
\end{figure*}

Preference-based RL (PBRL) operationalizes this concept: a teacher (human expert) provides preferences over the behaviors of two agents (or trajectories), which the agents then use to update their policies (see Figure~\ref{fig:fig1}). Compared to designing a reward function for each state-action pair, PBRL not only reduces the workload for human experts but also yields estimated rewards that are less biased and less susceptible to exploitation. Moreover, human preferences can guide agents to exhibit diverse styles, enabling better interpretability of why certain strategies are chosen.

However, current PBRL approaches face three key limitations: First, training instability arises from potential expert errors in preference judgments; Second, while experts provide trajectory-level preferences, most RL algorithms are optimized for dense reward models; Third, PBRL typically relies on pairwise trajectory preferences rather than leveraging inter-trajectory relationships to enhance learning.

To address these, we propose a novel PBRL baseline algorithm, ELO Rating-based RL (ERRL). For converting ordinal preferences to cardinal rewards, we adopt the ELO ranking system, widely used to measure player strength in zero-sum games. Specifically, we model human trajectory preferences as a zero-sum game, transforming preference signals into ELO ratings for model optimization. To handle sparse trajectory feedback, we employ maximum likelihood estimation (MLE), which yields a more stable reward decomposition than least squares methods. We also introduce engineering optimizations to enhance algorithm efficiency. Experiments on selected Atari games demonstrate that ERRL achieves performance comparable to state-of-the-art baselines. Ablation studies further validate the effectiveness of our approach. The code is available on \href{https://github.com/Zealoter/ERRL}{GitHub}.

%% file: p2preliminaries.tex
\subsection{Episodic RL with Trajectory Feedback}\label{subsec:ERLTF}
In classic RL settings, the research focuses on the Markov Decision Process (MDP) model, generally defined as a tuple: $\langle \mathcal{S}, \mathcal{A}, T, R \rangle$. Here, $\mathcal{S}$ and $\mathcal{A}$ denote the environment state space and agent action set, respectively. $T(s,a,s') = P(s'|s,a)$ represents the probability of transitioning to state $s'$ when action $a$ is taken in state $s$, while $R(s,a)$ denotes the value feedback for this state-action pair.

The goal of RL is to find a policy $\pi:~\mathcal{S}\rightarrow\mathcal{A}$ maximizing long term reward:
\begin{equation}
    \max G(s_t,a_t)=\max {\sum_{k=1}^{T}\gamma^k R(s_{t+k},a_{t+k})}\label{eq:rlgt}
\end{equation}
Where $\gamma$ is discount factor.

In real-world RL scenarios, specifying the reward function \( R(s,a) \) of the MDP model is often impractical; instead, only terminal outcomes of trajectories are observable. Consequently, RL agents frequently encounter challenges related to sparse or delayed feedback. A trajectory segment is defined as a sequence of state-action pairs \( \sigma = \left( (s_1, a_1), (s_2, a_2), \dots, (s_k, a_k) \right) \in (\mathcal{S} \times \mathcal{A})^k \). In response, trajectory-feedback RL models are typically formalized as a tuple: \( \langle \mathcal{S}, \mathcal{A}, T, \mathcal{R} \rangle \)~\cite{ref17}. Unlike standard MDPs, the environment provides only a cumulative evaluation \( \mathcal{R}(\sigma) \) over the entire trajectory \(\sigma\) upon termination.
\begin{equation}
    \mathcal{R}(\sigma)=\sum_{(s_i ,a_i )\in\sigma} R(s_i,a_i)
    \label{eq:sigmaR}
\end{equation}
To enhance training efficiency, it is essential to redistribute the cumulative trajectory reward to individual state-action pairs~\cite{ref15,ref17}, thereby transforming the trajectory-feedback RL problem into a standard MDP framework. This involves constructing a proxy reward function $\hat{r}(s,a)$ to approximate the state-action value, analogous to the reward function in traditional MDPs. In practice, we optimize this proxy reward function using the following loss function:
\begin{equation}
    \mathcal{L}(\hat{r})=\mathbb{E} _{\sigma \in \mathcal{D}}\left[ \left( \mathcal{R}(\sigma )-{\sum}_{(s_i ,a_i )\in\sigma  } \hat{r}(s_i ,a_i) \right) \right]
    \label{eq:LR1}
\end{equation}

\subsection{Ordinal Utility Theory and Cardinal Utility Theory}\label{subsec:OUCU}
Cardinal utility theory is a prominent concept in microeconomics during the 19th and early 20th centuries~\cite{ref18}. Its core idea is that utility can be measured and summed, with the basic unit of utility referred to as the utility unit. Utility units can be expressed numerically (e.g., $(1,2,3,\dots)$), analogous to how length is measured in meters. The mainstream RL reward function $R(s,a)$ mirrors the definition of cardinal utility.

Ordinal utility theory, proposed by Hicks and Allen in 1934~\cite{ref14}, posits that utility as a psychological phenomenon cannot be measured, as no reasonable unit for quantifying utility exists. They argue that consumers in the market do not measure the utility of goods but rather rank different goods.

Similarly, in certain RL problems, obtaining an accurate cardinal reward is highly challenging, whereas acquiring preferences over different trajectories is more feasible. For instance, chess typically involves 80–100 decision points, making it difficult even for professional chess players to estimate the numerical value of specific states. However, even beginners can judge the final match result.

\subsection{Elo Rating System}\label{subsec:elo-rating}
The ELO rating system~\cite{ref21} is an evaluation framework developed by Arpad Elo to measure player performance in zero-sum games. Widely adopted in competitive domains such as chess, Go, football, basketball, and esports, it serves as a standard for assessing player skill levels. Each player is assigned a latent rating score representing their average competitive ability: higher scores indicate stronger performance. Due to game randomness and player variability, observed outcomes may deviate from latent ratings. In the ELO model, this discrepancy is modeled as a fluctuating error term to reconcile actual performance with true ability. The core objective of ELO rating is to estimate player abilities (latent scores) as accurately as possible from win-loss outcomes. The formal ELO rating model is defined as follows:

Let the random variable $\xi$ follow a Logistic distribution, representing the gap between observed and latent scores:
\begin{equation}
    P(\xi \leq x) = F(x) = \frac{1}{1 + e^{-x}} \label{eq:eq1}
\end{equation}
Denote $R_A$ and $R_B$ as the current latent scores of Players A and B, respectively, and let $D = R_A - R_B$. The probability that Player A defeats Player B is:
\begin{equation}
    \begin{aligned}
        P(D) &= \frac{1}{2} + \int_{0}^{\frac{D}{2}} F'(x) \, dx \\
        &= \frac{1}{1 + e^{-\frac{D}{2}}}
    \end{aligned}
    \label{eq:eq2}
\end{equation}
To facilitate computation and intuitively reflect ability differences, ELO modifies the Logistic distribution's variance and sets the initial rating at 1500.

Assume the win rate of player A is $E(\sigma^A,\sigma^B)$:
\begin{equation}
    E(\sigma^A,\sigma^B)=\frac{1}{1+10^{(R_B-R_A)/400}}\label{eq:ea}
\end{equation}
$\mu$ denote result function, if $\mu(A)=1$ denote player A defeated player B, $\mu(A)=0$ denote player B defeated player A, Then player A's ELO rating is updated to:
\begin{equation}
    R_A^\prime = R_A+K(\mu(A)-E(\sigma^A,\sigma^B))
    \label{eq:eloupdate}
\end{equation}

\subsection{Performance Based RL (PBRL)}\label{subsec:shp}
PBRL~\cite{ref16} is a typical RL algorithm leveraging ordinal utility, which partitions each trajectory into fixed-length $k$ sub-trajectories, with $\sigma^A \succ \sigma^B$ denoting human experts' preference for sub-trajectory $\sigma^A$ over $\sigma^B$.

If experts judge $\sigma^A \succ \sigma^B$, it implies a high probability that $\mathcal{R}(\sigma^A) > \mathcal{R}(\sigma^B)$:
\begin{equation}
    \begin{aligned}
    & R(s^A_1, a^A_1) + R(s^A_2, a^A_2) + \dots + R(s^A_{k_A}, a^A_{k_A}) \notag \\
    &> R(s^B_1, a^B_1) + R(s^B_2, a^B_2) + \dots + R(s^B_{k_B}, a^B_{k_B})
    \end{aligned}
\end{equation}
Thus, under the proxy reward function $\hat{r}$, it is reasonable to assume the probability of an expert preferring $\sigma^A \succ \sigma^B$ is:
\begin{equation}
    \hat P(\sigma^A\succ\sigma^B)=\frac{\exp \sum\limits_{i=1}^{i=k_A} \hat{r}(s^A_i, a^A_i)}{\exp \sum\limits_{i=1}^{i=k_A} \hat{r}(s^A_i, a^A_i)+\exp \sum\limits_{i=1}^{i=k_B} \hat{r}(s^B_i, a^B_i)}
    \label{eq:p_ab}
\end{equation}
Under this framework, expert preference results $\left( \sigma^A, \sigma^B, \mu \right)$ are stored in database $\mathcal{D}$, where $\mu$ denotes the preference indicator: $\mu(A) = 1$ indicates $\sigma^A \succ \sigma^B$, and $\mu(A) = 0$ indicates $\sigma^B \succ \sigma^A$. Leveraging these preferences, we optimize the proxy reward function $\hat{r}$ using cross-entropy loss and train the RL agent via the proxy reward function $\hat{r}$:
\begin{equation}
    \mathcal{L}(\hat{r})=-\sum_{( \sigma^A,\sigma^B\,\mu)\in \mathcal{D}}
    \left(\begin{aligned}
    \mu(A)\log \hat{P}(\sigma^A\succ\sigma^B)+ \\ \mu(B)\log \hat{P}(\sigma^B\succ\sigma^A)
    \end{aligned}\right)
    \label{eq:LR2}
\end{equation}

%% file: p4methodology.tex
In this section, we introduce our novel method, ELO-Rating based RL (ERRL), which replaces numerical rewards with trajectory rankings via ELO ratings and employs a simplified return decomposition to assign ELO-based values to state-action pairs.

\subsection{Trajectory Reward Estimation via ELO Rating}\label{subsec:estimation-trail-reward-via-elo-Rating}

Given two trajectories $\sigma^A$ and $\sigma^B$, suppose $\sigma^A$ outperforms $\sigma^B$ (denoted $\sigma^A \succ \sigma^B$, e.g., Player A wins in chess). Let the trajectory reward be $\mathcal{R}(\sigma^X) = \overline{\mathcal{R}(\sigma^X)} + \epsilon$, where $\overline{\mathcal{R}(\sigma^X)}$ is the expected reward of trajectory $\sigma^X$ and $\epsilon$ models reward fluctuation. Estimating $\overline{\mathcal{R}(\sigma^X)}$ is challenging: it either requires exhaustive exploration of the solution space (prohibitive in computation) or human experts to design intermediate rewards for efficient guidance.

While assigning precise rewards $\mathcal{R}(\sigma^X)$ is difficult, comparing trajectories is more feasible. Leveraging ordinal utility theory, we define $\mathcal{U}(\sigma)$ as the ELO rating of trajectory $\sigma$, using $\mathcal{U}(\sigma)$ as the trajectory reward for RL training. When $\sigma^A \succ \sigma^B$, this corresponds to "player" $\sigma^A$ defeating "player" $\sigma^B$ (with score $S_A = 1$); if experts cannot distinguish between trajectories ($\sigma^A \approx \sigma^B$), this is treated as a draw ($S_A = 0.5$). The event $\sigma^A \succ \sigma^B$ implies a high probability that $\hat{\mathcal{U}}(\sigma^A) > \hat{\mathcal{U}}(\sigma^B)$, allowing the ELO rating $\hat{\mathcal{U}}(\sigma^A)$ to be updated as:
\begin{equation}
    E(\sigma^A,\sigma^B)=\frac{1}{1+10^{(\hat{\mathcal{U}}(\sigma^B)-\hat{\mathcal{U}}(\sigma^A))/\eta}}
    \label{eq:ea2}
\end{equation}
\begin{equation}
    \hat{\mathcal{U}}^{\prime}(\sigma^A)=\hat{\mathcal{U}}(\sigma^A)+K(S_A-E(\sigma^A,\sigma^B))
    \label{eq:hatU}
\end{equation}
Where $\eta$ and $K$ are the parameters to adjust the training fluctuation.

\subsection{Indirect Redistribution of Long-Term Reward}\label{subsec:indirect-redistribution-of-long-term-reward}
$\hat{\mathcal{U}}(\sigma)$ is estimating on the entire trajectory,
which should be the sum of the ELO rating for all state-action pairs:
\begin{equation}
    \hat{\mathcal{U}}(\sigma^X)=\sum_{t=1}^{k_X} \hat U(s_t^X,a_t^X)
    \label{eq:r_elo}
\end{equation}
In value-based RL, the value function inherently encodes a one-step reward estimate \( R(s,a) = G(s,a) - \gamma G(s',a') \), obviating the need for a proxy reward function. We employ a deep neural network \( \hat{G}_{\text{elo}}(s,a) \) to estimate the long-term ELO rating for state-action pair \( (s,a) \). The score for the entire trajectory \( \sigma^X \) is denoted as \( \hat{\mathcal{U}}(\sigma^X) \):
\begin{equation}
    \begin{aligned}
        \hat{\mathcal{U}}(\sigma^X)&=\sum_{t=1}^{k_X} \hat U(s_t^X,a_t^X)\\
        &=\sum_{t=1}^{k_X} \left[ \hat G_{\text{elo}}(s^X_t,a^X_t)-\gamma \hat G_{\text{elo}}(s^X_{t+1},a^X_{t+1}) \right]
    \end{aligned}
    \label{eq:g_elo}
\end{equation}
The long-term reward for each state-action pair is updated as (proved in Appendix~\ref{sec:appendix-a}):
\begin{equation}
    \hat {G}^\prime_{\text{elo}}(s^X_t,a^X_t)= \hat G_{\text{elo}}(s^X_t,a^X_t)+\frac{K(S_X-E_X)}{k_X}
    \label{eq:Gprimeelo}
\end{equation}

\subsection{Combining RL and Elo Rating}\label{subsec:ac-rl-with-elo-Rating}
At the implementation level, our algorithm is compatible with mainstream RL frameworks. It can be seamlessly integrated into any RL algorithm by modifying the reward function alone. In Algorithm~\ref{alg:algorithm}, we demonstrate this compatibility by integrating Elo rating with an actor-critic RL framework.

\begin{algorithm}
    \caption{AC RL with Elo Rating}
    \begin{algorithmic}[1]
        \State Initialize $\mathcal{D}\leftarrow \varnothing $
        \For{$\ell = 1,2,\dots$}
            \State Collect a rollout trajectory $\sigma$ using the current policy, and store trajectory $\sigma$ into the replayer buffer $\mathcal{D}$.
            \State Sample $N$ trajectories $W=\{\sigma^i\in \mathcal{D}\}^N_{i=1}$ from the replay buffer.
            \For{$i=1,2,\dots,N$}
                \State Generates a shuffled integer list $Y$ from 1 to $N$.
                \State Use equation$~$\ref{eq:ea2} to get a new long term Elo rating estimation.
                \begin{center}
                    $E(\sigma^i,\sigma^{Y[i]})=1/(1+10^{(\hat{\mathcal{U}}(\sigma^{Y[i]})-\hat{\mathcal{U}}(\sigma^i))/\eta})$
                \end{center}
            \EndFor
        \State Use equation$~$\ref{eq:Gprimeelo} to gradient update on the critic network.
        \begin{equation}
            \mathcal{L}(\hat {G}^\prime_\text{elo})=\sum_{i=1}^{N}\sum_{(s^i,a^i)\in \sigma^i}\left(\frac{K(\mu(\sigma^i)-E(\sigma^i,\sigma^{Y[i]}))}{k_i} \right)
            \label{eq:LGElo}
        \end{equation}
        \State Use critic network to update actor network.
        \EndFor
    \end{algorithmic}
    \label{alg:algorithm}
\end{algorithm}

Notably, expert preferences are expected to significantly influence the outcomes of ordinal-based RL. Theoretically, the ELO rating parameters \( K \) and \( \eta \) will also impact performance, which we will analyze in detail in Section~\ref{sec:experiments}.

%% file: p5experiments.tex
In this section, we conduct experiments on a suite of Atari benchmark tasks to evaluate the empirical performance of our proposed method. We compare ERRL with several state-of-the-art baselines and perform ablation studies to analyze its components.

\subsection{Experimental Setups}
To assess ERRL's behavior, we compare its performance against multiple baseline algorithms across four Atari tasks, all featuring long-term feedback (up to 5000 steps). In these ordinal-based RL settings, trajectories record key events for expert preference judgments rather than returning numerical rewards. We use the open-source RL framework RLLib~\cite{ref24} as the baseline, which adopts DeepMind's standard preprocessing pipeline~\cite{ref7} and enables efficient parallel training. All results are fine-tuned, with detailed configurations provided in Appendix~\ref{sec:appendix-b}.

As our experiments rely on ordinal RL, generating the ordinal preference model is critical for learning outcomes. We illustrate the ordinal generation process using Pong as a case study.

Pong simulates table tennis, where scoring occurs when the ball lands beyond the opponent's baseline (+1 point) or the player fails to return a shot (-1 point). The first player to reach 21 points wins. The final trajectory reward is defined as $\mathcal{R}(\sigma) = \text{self\_score} - \text{opponent\_score}$. Our evaluation criterion for two trajectories $\sigma^A$ and $\sigma^B$ is:

\begin{enumerate}
    \item If $\mathcal{R}(\sigma^A) > \mathcal{R}(\sigma^B)$ or $\mathcal{R}(\sigma^A) < \mathcal{R}(\sigma^B)$, then $\sigma^A \succ \sigma^B$ or $\sigma^A \prec \sigma^B$; otherwise, proceed to the next criterion.
    \item If both $\mathcal{R}(\sigma^A) > 0$ and $\mathcal{R}(\sigma^B) > 0$, then $\sigma^A \approx \sigma^B$; otherwise, proceed to the third criterion.
    \item If the trajectory length $k_A > k_B$ or $k_A < k_B$, then $\sigma^A \succ \sigma^B$ or $\sigma^A \prec \sigma^B$; otherwise, $\sigma^A \approx \sigma^B$.
\end{enumerate}
Here, $k$ denotes the trajectory length, and the game terminates automatically after 5000 steps to prevent excessively long trajectories.

This design follows two principles: when direct victory is unfeasible, the strategy shifts to prolonged play, aiming to maximize ball returns; when victory is achievable, any means to secure the win are prioritized. No additional constraints are imposed. For detailed game configurations, refer to Appendix~\ref{sec:appendix-b}.

\subsection{Performance in Atari Benchmark}
\begin{figure*}
    \centering
    \includegraphics[width=\textwidth]{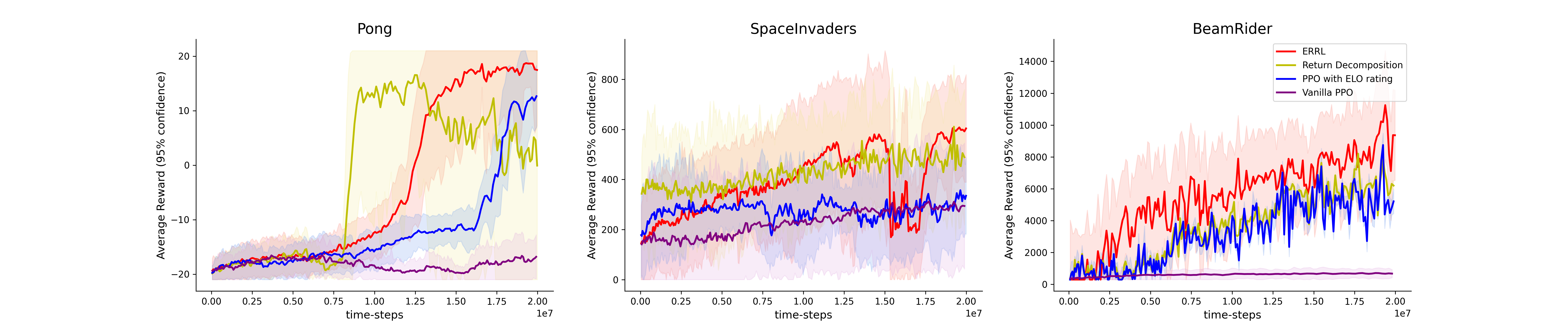}
    \caption{
        Learning curves on a suite of Atari benchmark tasks with trajectory feedback.
        The shaded region indicates 95\% confidence interval.
        Due to the trajectory length is indeterminate,
        we set up an evaluation point every 60 trajectories, about $3\times10^{4}\sim 3\times10^{5}$ time-steps.
    }\label{fig:fig2}
\end{figure*}

The overall performance comparison in Figure~\ref{fig:fig2} demonstrates that ERRL consistently outperforms baseline algorithms across these tasks, achieving higher sample efficiency and converging to superior policies. Due to the sparse nature of rewards, vanilla PPO struggles to extract meaningful information. Existing reward decomposition methods, designed for numerical rewards, introduce instability when applied to ordinal preferences. In contrast, ERRL leverages ELO ratings to convert ordinal feedback into scalable rewards, amplifying performance differences between trajectories. This approach proves particularly effective in long-horizon RL problems, where ELO ratings more effectively highlight trajectory quality than raw numerical rewards, enabling the agent to identify optimal strategies.

\subsection{Ablation Study}\label{subsec:ablation-study}
We study the impact of $\eta$ in the ERRL algorithm under a fixed learning rate of $1 \times 10^{-4}$. For the parameter $K$, we follow the original ELO design by setting $K = 0.04 \times \eta$.

As discussed in Section~\ref{sec:methodology}, $\eta$ regulates the fluctuation range of ELO ratings: smaller $\eta$ introduces higher zero-sum game randomness, leading to larger training volatility but faster convergence; larger $\eta$ reduces randomness, decreasing training fluctuations but slowing down learning. Thus, an optimal $\eta$ is needed to balance training speed and stability.
\begin{figure*}
    \centering
    \includegraphics[width=\textwidth]{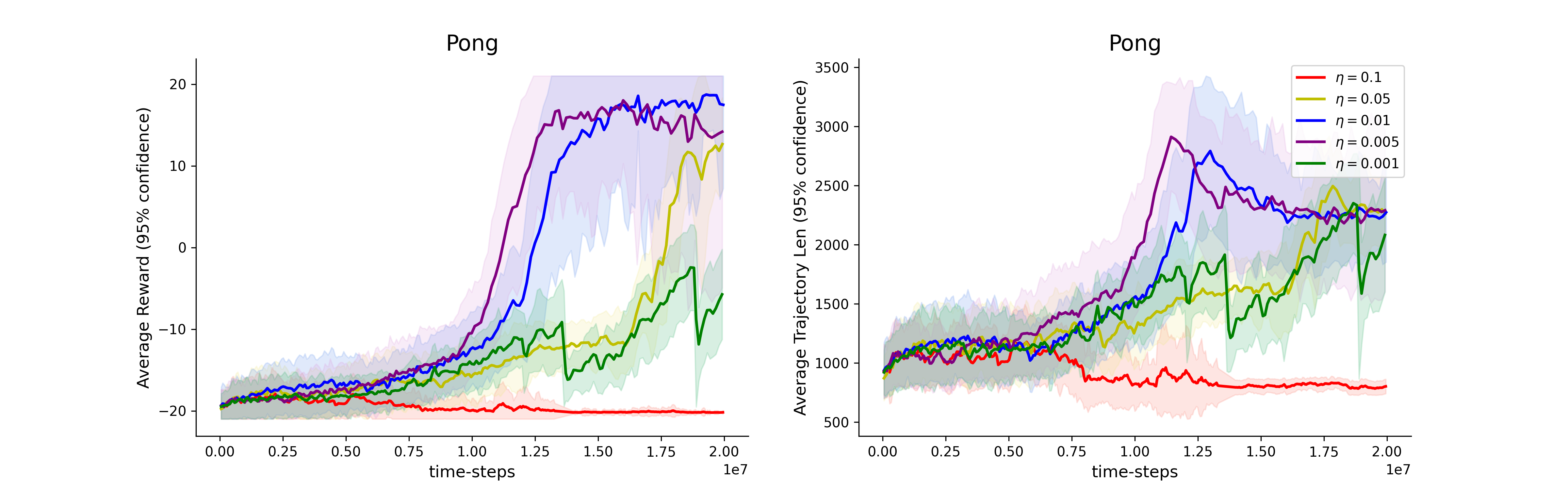}
    \caption{
        Learning curves of ERRL in Pong with different choices of hyper-parameter $\eta$.
        The curves with $\eta=0.01$ correspond to the default implementation.
    }\label{fig:fig3}
\end{figure*}

We conduct an ablation study on the hyperparameter $\eta$, which controls the fluctuation range of the ELO rating. Given that game time steps vary during gameplay, we link $\eta$ to the trajectory length, such that the effective $\eta$ value used in ELO rating calculations is $\eta_{\text{elo}}=\eta_{\text{our}}\times\overline{k_X}$. For simplicity, we use $\eta$ to denote $\eta_{\text{our}}$ throughout the remainder of the text unless otherwise specified.

We evaluate ERRL with $\eta \in \{0.1, 0.05, 0.01, 0.005, 0.001\}$. As illustrated in Figure~\ref{fig:fig3}, this hyperparameter significantly impacts RL training. Larger $\eta$ values lead to excessive training volatility and potential model collapse, whereas smaller $\eta$ values slow down convergence and increase the risk of overfitting. An optimal range for $\eta$ is identified as $[0.05, 0.005]$.

Across all trajectories, regardless of $\eta$'s magnitude, the final time steps converge within the range of 2100–2400. This indicates that, under consistent expert preferences, the choice of $\eta$ does not alter the learned optimal strategies. We further discuss the relationship between strategy selection and expert preferences in Appendix~\ref{sec:appendix-c}.

%% file: p3related_work.tex
\textbf{RL without Environmental Reward}.

While traditional RL relies on environmental rewards, recent studies have explored reward-free RL frameworks, which can be categorized into two main approaches. The first approach focuses on extracting discriminative information from trajectories to define surrogate rewards. Eysenbach \textit{et al.}~\cite{ref28} proposed a pre-training method where a discriminator is trained to identify the agent that generated a trajectory. The agent aims to produce trajectories that are easily distinguishable, driving it to seek unique states. Hu \textit{et al.}~\cite{ref26} uses multidimensional features to determine the Pareto frontier, with only frontier agents proceeding to the next training phase. This approach enhances agent diversity but is typically limited to diversity-driven tasks. The second approach integrates human preferences into RL, such as Preference-Based RL (PBRL)~\cite{ref16} discussed in Section~\ref{subsec:shp}. Additionally, Kimin \textit{et al.}~\cite{ref13} proposed PEBBLE, which has shown remarkable performance in reward-free settings.

\textbf{Sparse Rewards RL}.

Sparse rewards remain a critical challenge for RL applications~\cite{ref11, ref8}, with research addressing this through multiple strategies. A direct solution is to incorporate human expertise~\cite{ref1, ref4}, which can yield successful results if expert knowledge is well-codified. However, heavy reliance on human input is costly and may inadvertently constrain agent capabilities~\cite{ref1, ref4, ref5}. Another approach is reward decomposition, which distributes cumulative rewards to individual state-action pairs~\cite{ref15, ref20}. This often requires reward shaping to achieve effectiveness. Marcin \textit{et al.}~\cite{ref10} leverages historical information reuse for sparse reward tasks, though this is limited to domains with simple, well-defined states.

%% file: p7summary_and_conclusion.tex
We propose a novel preference-based RL algorithm, ERRL (ELO-Rating based Reinforcement Learning), which demonstrates superior performance in 5000-step long-term trajectory-feedback RL tasks. ERRL eschews classical numerical rewards, instead estimating ordinal rewards for each trajectory via expert preferences and the ELO rating system. Additionally, a new reward decomposition method is introduced to mitigate training volatility under the ERRL framework. Experimental results show that ERRL outperforms state-of-the-art methods in solution quality. For future work, we will explore the theoretical foundations of integrating ELO rating with RL, conduct more detailed ablation studies, investigate algorithm effectiveness under ambiguous expert preferences, and explore the feasibility of ELO rating in extended domains including reward shaping~\cite{ref27}, games with complex payoffs (e.g., Texas hold'em and Hanabi)~\cite{ref25}, and research on RL agent diversity~\cite{ref26, ref28}.

%% file: p91Appendix_A.tex
$\hat{\mathcal{U}}(\sigma^A)$ is the ELO rating estimate of the entire trajectory $\sigma^A$,
in trajectory $\sigma^A$, let the long-term ELO rating $\hat U(s_t^X,a_t^X)$ for each state-action pair update as $\delta_t^X$,
\begin{equation}
    \hat {U}^\prime(s_t^A,a_t^A)=\hat U(s_t^A,a_t^A)+\delta_t^A
    \label{eq:a1}
\end{equation}
In theory, any update can be regarded as an effective update range as long as it is randomly sampled so that the $\delta_t^X$ distribution satisfies the $K(S_X-E_X)=\delta_t^X$ constraint.
But this obviously leads to too large variance,
which is not suitable for iterative calculation.
So we might as well consider $\delta_t^A$ as the assumption of the ELO rating on the player's ability to fluctuate,
and assume that the estimation error for each state-action pair is independent and uniformly obeys the Logistic distribution:
\begin{equation}
    \delta_t^A\sim\text{Logistic}(x)=\frac{e^{-x}}{(1+e^{-x})^2}
    \label{eq:a2}
\end{equation}
According to maximum likelihood estimation,
the above problem can be transformed into an optimization problem.
The current formula $~$\ref{eq:a3} takes the maximum value of $x_t$ as the most likely update range:
\begin{equation}
    \begin{split}
        \max \prod_{t=1}^{k_A}\frac{e^{-\delta_t^A}}{(1+e^{-\delta_t^A})^2}\\
        \text{s.t.}~~~~(S_A-E_A)=\sum_{t=1}^{k_A}\delta_t^A
    \end{split}
    \label{eq:a3}
\end{equation}
Taking the logarithm of the original formula$~$\ref{eq:a3}, there are:
\begin{equation}
    \max-\sum_{t=1}^{k_A}\left( \delta_{t}^{A}+2\ln(1+e^{-\delta_{t}^A}) \right)
    \label{eq:a4}
\end{equation}
Bringing into $\sum_{t=1}^{k_A}=(S_A-E_A)$
\begin{equation}
    \begin{aligned}
        &~~~\sum_{t=1}^{k_A}\left( \ln(1+e^{-\delta_{t}^A}) \right) \\
        &=\sum_{t=1}^{k_A}\left( \frac{\ln(1+e^{\delta_{t}^A})}{e^{\delta_{t}^A}} \right)
    \end{aligned}
    \label{eq:a5}
\end{equation}
Finally:
\begin{equation}
    \begin{split}
        \min \sum_{t=1}^{k_A}\ln(1+e^{\delta^A_t})\\
        \text{s.t.}~~~~K(S_A-E_A)=\sum_{t=1}^{k_A}\delta_t^A
    \end{split}
    \label{eq:a6}
\end{equation}

\begin{equation}
    \begin{split}
        \min \sum_{t=1}^{k_A}\ln(1+e^{\delta^A_t})\\
        \text{s.t.}~~~~K(S_A-E_A)=\sum_{t=1}^{k_A}\delta_t^A
    \end{split}
    \label{eq:a7}
\end{equation}
It is easy to know that when $\delta_t^A=\frac{K(S_A-E_A)}{k_A}$,
the cost function obtains the minimum value,
so the new model is updated as:
\begin{equation}
    \hat {G}_{\text{elo}}^\prime(s_t^A,a_t^A)=\gamma\hat {G}_{\text{elo}}(s_{t+1}^A,a_{t+1}^A)+U(s_t^A,a_t^A)+\delta_t^A \\
    \label{eq:a8}
\end{equation}
Bringing into $U(s_t^A,a_t^A)=\hat{G}_{\text{elo}}(s_{t}^A,a_{t}^A)-\gamma\hat{G}_{\text{elo}}(s_{t+1}^A,a_{t+1}^A)$,
\begin{equation}
    \hat {G}_{\text{elo}}^\prime(s_t^A,a_t^A)=\hat{G}_{\text{elo}}(s_{t}^A,a_{t}^A)+\delta_t^A
    \label{eq:a9}
\end{equation}

%% file: p92Appendix_B.tex
\subsection*{Hyper-Parameter Configuration For Atari}\label{subsec:B1}
In terms of hardware configuration, the CPU is AMD thread-rippers 3990WX with 32 cores and 64 threads,
the graphics card is Nvidia RTX 3080Ti,
and the memory is 128GB .
During training, 60 workers are used to collect data at the same time to improve the experimental efficiency.
The training method of agent is PPO .
We refer to the default settings of PPO~\cite{ref24,ref22},
the main hyperparameter settings are shown in the figure~\ref{tab:Hyper-Parameter}.

\begin{table}[h]
    \caption{Hyper-Parameter}
    \label{tab:Hyper-Parameter}
    \centering
    \begin{tabular}{ll}
        \toprule
        Hyper-Parameter                     & Default Configuration \\
        \midrule
        discount factor $\gamma$            & 0.99                  \\
        activation                          & ReLU                  \\
        stacked frames in agent observation & 4                     \\
        state mode                          & memory                \\
        actor agent num                     & 60                    \\
        \midrule
        network architecture                & PPO                   \\
        optimizer                           & Adam                  \\
        learning rate                       & $1\cdot10^{-4}$       \\
        gradient steps per environment step & 30000$\sim$300000     \\
        \midrule
        ELO $\eta$                          & 0.01                  \\
        ELO $K$                             & $0.04\times\eta$        \\
        \bottomrule
    \end{tabular}
\end{table}

Our open source main experimental code and details can be found in \href{https://github.com/Zealoter/ERRL}{Github},
as our experiments are still ongoing, we will gradually improve our code and experimental details.
\subsection*{Expert Preferences}\label{subsec:B2}
\subsubsection*{Pong}\label{subsubsec:B21}
In Pong game,
you control the right paddle,
you compete against the left paddle controlled by the computer.
Unlike reality ping-pong,
the upper and lower sides of the table have boundaries,
regardless of the ball going out of bounds.
Evaluation criteria for trajectory $\sigma^A$ and trajectory $\sigma^B$ are:
\begin{enumerate}
    \item If $\mathcal{R}(\sigma^A)>\mathcal{R}(\sigma^B)$ or $\mathcal{R}(\sigma^A)<\mathcal{R}(\sigma^B)$ then $\sigma^A\succ\sigma^B$ or $\sigma^A\prec\sigma^B$, otherwise see the second item.
    \item If $\mathcal{R}(\sigma^A)>0$ and $\mathcal{R}(\sigma^B)>0$ then $\sigma^A\approx\sigma^B$, otherwise see the third item.
    \item If $k_A>k_B$ or $k_A<k_B$ then $\sigma^A\succ\sigma^B$ or $\sigma^A\prec\sigma^B$, otherwise the final result is $\sigma^A\approx\sigma^B$.
\end{enumerate}
\subsubsection*{SpaceInvaders}\label{subsubsec:B22}
In this game,
your goal is to fire laser cannons at space invaders to destroy them before they reach Earth.
The game is over when you lose all your lives under enemy's fire,
or when they reach Earth.Since the aliens will gradually approach the earth unless they are eliminated.
Obviously,
the score is not as important as the time in this game.
This is because the agent may give up killing the front row in order to kill the high-scoring aliens in the back row.
So the evaluation criteria for trajectory $\sigma^A$ and trajectory $\sigma^B$ are:
\begin{enumerate}
    \item If $K_A>K_B$ or $K_A<K_B$ then $\sigma^A\succ\sigma^B$ or $\sigma^A\prec\sigma^B$, otherwise see the second item.
    \item If $\mathcal{R}(\sigma^A)>\mathcal{R}(\sigma^B)$ or $\mathcal{R}(\sigma^A)<\mathcal{R}(\sigma^B)$ then $\sigma^A\succ\sigma^B$ or  $\sigma^A\prec\sigma^B$, otherwise the final result is $\sigma^A\approx\sigma^B$.
\end{enumerate}

\subsubsection*{BeamRider}\label{subsubsec:B23}
In this game,
you control a space-ship that travels forward at a constant speed.
You can only steer it sideways between discrete positions.
However, at the end of the level only torpedoes can defeat the boss,
and the number of torpedoes is limited,
so the preferences of this game are difficult to design.
From the player's point of view it's always beneficial to live longer,
so we combined time-steps length and score to design BeamRider's preferences.
The evaluation criteria for trajectory $\sigma^A$ and trajectory $\sigma^B$ are:
\begin{enumerate}
    \item If $\mathcal{R}(\sigma^A)+K_A>\mathcal{R}(\sigma^B)+K_B$ or $\mathcal{R}(\sigma^A)+K_A<\mathcal{R}(\sigma^B)+K_B$ then $\sigma^A\succ\sigma^B$ or $\sigma^A\prec\sigma^B$, otherwise the final result is $\sigma^A\approx\sigma^B$.

\end{enumerate}

%% file: p93Appendix_C.tex
Here we discuss several settings of expert preference using Pong and
explore the impact of different expert preferences on training.

\textbf{Preference 1: normal mode}

This mode is the same as the previous.
\begin{enumerate}
    \item If $\mathcal{R}(\sigma^A)>\mathcal{R}(\sigma^B)$ or $\mathcal{R}(\sigma^A)<\mathcal{R}(\sigma^B)$ then $\sigma^A\succ\sigma^B$ or $\sigma^A\prec\sigma^B$, otherwise see the second item.
    \item If $\mathcal{R}(\sigma^A)>0$ and $\mathcal{R}(\sigma^B)>0$ then $\sigma^A\approx\sigma^B$, otherwise see the third item.
    \item If $k_A>k_B$ or $k_A<k_B$ then $\sigma^A\succ\sigma^B$ or $\sigma^A\prec\sigma^B$, otherwise the final result is $\sigma^A\approx\sigma^B$.
\end{enumerate}

\textbf{Preference 2: reward mode}


\begin{figure}[htbp]
    \centering
    \includegraphics[width=\textwidth]{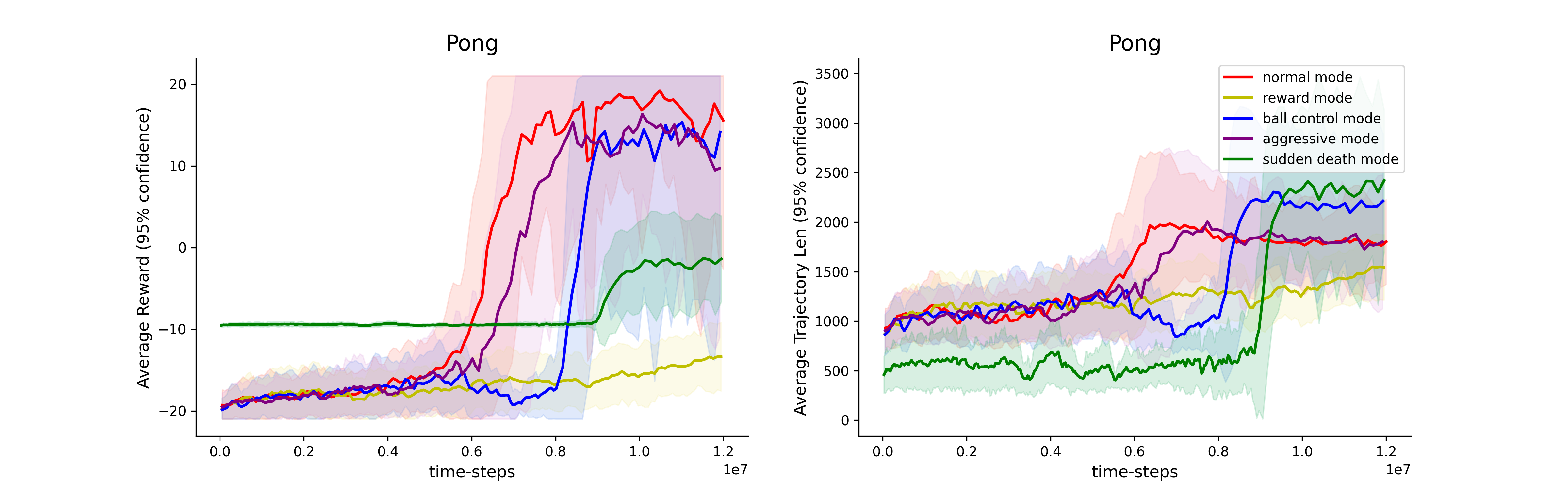}
    \caption{
        It can be seen that the preference has a great influence on ERRL,
        and changing the preference may change the final policy of the agent.
    }
    \label{fig:fig4}
\end{figure}

In this mode, preference comes only from reward.
\begin{enumerate}
    \item If $\mathcal{R}(\sigma^A)>\mathcal{R}(\sigma^B)$ or $\mathcal{R}(\sigma^A)<\mathcal{R}(\sigma^B)$ then $\sigma^A\succ\sigma^B$ or $\sigma^A\prec\sigma^B$, otherwise the final result is $\sigma^A\approx\sigma^B$.
\end{enumerate}

\textbf{Preference 3: ball control mode}

In this mode,
trajectories with more time-steps are always better in the same reward.
That's why it's called ball control mode.

\begin{enumerate}
    \item If $\mathcal{R}(\sigma^A)>\mathcal{R}(\sigma^B)$ or $\mathcal{R}(\sigma^A)<\mathcal{R}(\sigma^B)$ then $\sigma^A\succ\sigma^B$ or $\sigma^A\prec\sigma^B$, otherwise see the second item.
    \item If $k_A>k_B$ or $k_A<k_B$ then $\sigma^A\succ\sigma^B$ or $\sigma^A\prec\sigma^B$, otherwise the final result is $\sigma^A\approx\sigma^B$.
\end{enumerate}

\textbf{Preference 4: aggressive mode}

We want the agent to beat the opponent as quickly as possible,
so when agent is able to beat the opponent,
less time-steps is the better.
\begin{enumerate}
    \item If $\mathcal{R}(\sigma^A)>\mathcal{R}(\sigma^B)$ or $\mathcal{R}(\sigma^A)<\mathcal{R}(\sigma^B)$ then $\sigma^A\succ\sigma^B$ or $\sigma^A\prec\sigma^B$, otherwise see the second item.
    \item If $\mathcal{R}(\sigma^A)>0$ and $\mathcal{R}(\sigma^B)>0$ then see the third item, if $\mathcal{R}(\sigma^A)<0$ and $\mathcal{R}(\sigma^B)<0$ then see the fourth item, $\sigma^A\approx\sigma^B$.
    \item If $k_A<k_B$ or $k_A>k_B$ then $\sigma^A\succ\sigma^B$ or $\sigma^A\prec\sigma^B$, otherwise the final result is $\sigma^A\approx\sigma^B$.
    \item If $k_A>k_B$ or $k_A<k_B$ then $\sigma^A\succ\sigma^B$ or $\sigma^A\prec\sigma^B$, otherwise the final result is $\sigma^A\approx\sigma^B$.
\end{enumerate}

\textbf{Preference 5: sudden death mode}

In this mode,
if $\mathcal{R}(\sigma^X)\le -5$,
game over directly,
and the rest is the same as preference 1.
\begin{enumerate}
    \item If $\mathcal{R}(\sigma^A)>\mathcal{R}(\sigma^B)$ or $\mathcal{R}(\sigma^A)<\mathcal{R}(\sigma^B)$ then $\sigma^A\succ\sigma^B$ or $\sigma^A\prec\sigma^B$, otherwise see the second item.
    \item If $\mathcal{R}(\sigma^A)>0$ and $\mathcal{R}(\sigma^B)>0$ then $\sigma^A\approx\sigma^B$, otherwise see the third item.
    \item If $k_A>k_B$ or $k_A<k_B$ then $\sigma^A\succ\sigma^B$ or $\sigma^A\prec\sigma^B$, otherwise the final result is $\sigma^A\approx\sigma^B$.
\end{enumerate}

It can be seen that inappropriate preferences may fail to train useful agents.
In the case of ball control mode,
the agent gets a policy that can play more time-steps in the game.
However, the rest of the models did not reach satisfactory conclusions,
so we will explore the relationship between expert preference and final strategy in future research.

%% file: ERRL.bbl
\begin{thebibliography}{99}
    \small
    \bibitem{ref1} Oh J, Guo Y, Singh S, et al.\ Self-imitation learning[C].\ International Conference on Machine Learning.\ PMLR, 2018: 3878-3887.

    \bibitem{ref2} Silver D, Hubert T, Schrittwieser J, et al.\ A general reinforcement learning algorithm that masters chess, shogi, and Go through self-play[J].\ Science, 2018, 362(6419): 1140-1144.

    \bibitem{ref3} Minh V, Kavukcuoglu K, Silver D, et al.\ Human-level control through deep reinforcement learning[J].\ nature, 2015, 518(7540): 529-533.

    \bibitem{ref4} Vinyls O, Babushka I, Czarnecki W M, et al.\ Grandmaster level in StarCraft II using multi-agent reinforcement learning[J].\\ Nature, 2019, 575(7782): 350-354.

    \bibitem{ref5} Silver D, Schrittwieser J, Simonyan K, et al.\ Mastering the game of go without human knowledge[J].\ nature, 2017, 550(7676): 354-359.

    \bibitem{ref6} Berner C, Brockman G, Chan B, et al.\ Dota 2 with large scale deep reinforcement learning[J].\ arXiv preprint arXiv:1912.06680, 2019.

    \bibitem{ref7} Mnih V, Kavukcuoglu K, Silver D, et al.\ Playing atari with deep reinforcement learning[J].\ arXiv preprint arXiv:1312.5602, 2013.

    \bibitem{ref8} Nair V, Bartunov S, Gimeno F, et al.\ Solving mixed integer programs using neural networks[J].\ arXiv preprint arXiv:2012.13349, 2020.

    \bibitem{ref9} Senior A W, Evans R, Jumper J, et al.\ Improved protein structure prediction using potentials from deep learning[J].\ Nature, 2020, 577(7792): 706-710.

    \bibitem{ref10} Andrychowicz M, Wolski F, Ray A, et al.\ Hindsight experience replay[J].\ Advances in neural information processing systems, 2017, 30.

    \bibitem{ref11} Degrave J, Felici F, Buchli J, et al.\ Magnetic control of tokamak plasmas through deep reinforcement learning[J].\ Nature, 2022, 602(7897): 414-419.

    \bibitem{ref12} Brown N, Sandholm T. Superhuman AI for multiplayer poker[J].\ Science, 2019, 365(6456): 885-890.

    \bibitem{ref13} Lee K, Smith L, Abbeel P. Pebble: Feedback-efficient interactive reinforcement learning via relabeling experience and unsupervised pre-training[J].\ arXiv preprint arXiv:2106.05091, 2021.

    \bibitem{ref14} Hicks J R, Allen R G D. A reconsideration of the theory of value.\ Part I[J].\ Economica, 1934, 1(1): 52-76.

    \bibitem{ref15} Ren Z, Guo R, Zhou Y, et al.\ Learning Long-Term Reward Redistribution via Randomized Return Decomposition[J].\ arXiv preprint arXiv:2111.13485, 2021.

    \bibitem{ref16} Christiano P F, Leike J, Brown T, et al.\ Deep reinforcement learning from human preferences[J].\ Advances in neural information processing systems, 2017, 30.

    \bibitem{ref17} Efroni Y, Merlis N, Mannor S. Reinforcement learning with trajectory feedback[J].\ arXiv preprint arXiv:2008.06036, 2020.

    \bibitem{ref18} Stigler G J. The development of utility theory.\ I[J].\ Journal of political economy, 1950, 58(4): 307-327.

    \bibitem{ref19} Lee K, Smith L, Dragan A, et al.\ B-Pref: Benchmarking Preference-Based Reinforcement Learning[J].\ arXiv preprint arXiv:2111.03026, 2021.

    \bibitem{ref20} Raposo D, Ritter S, Santoro A, et al.\ Synthetic returns for long-term credit assignment[J].\ arXiv preprint arXiv:2102.12425, 2021.

    \bibitem{ref21} Elo, Arpad E , The Proposed USCF Rating System, Its Development, Theory, and Applications[J].\ Chess Life, 1968 XXII (8): 242–247.

    \bibitem{ref22} Schulman J, Wolski F, Dhariwal P, et al.\ Proximal policy optimization algorithms[J].\ arXiv preprint arXiv:1707.06347, 2017.

    \bibitem{ref23} Schulman J, Moritz P, Levine S, et al.\ High-dimensional continuous control using generalized advantage estimation[J].\ arXiv preprint arXiv:1506.02438, 2015.

    \bibitem{ref24} Liang E, Liaw R, Nishihara R, et al.\ RLlib: Abstractions for distributed reinforcement learning[C].\ International Conference on Machine Learning.\ PMLR, 2018: 3053-3062.

    \bibitem{ref25} Bard N, Foerster J N, Chandar S, et al.\ The hanabi challenge: A new frontier for ai research[J].\ Artificial Intelligence, 2020, 280: 103216.

    \bibitem{ref26} Hu Y, Wang W, Jia H, et al.\ Learning to utilize shaping rewards: A new approach of reward shaping[J].\ Advances in Neural Information Processing Systems, 2020, 33: 15931-15941.

    \bibitem{ref27} Ng A Y, Harada D, Russell S. Policy invariance under reward transformations: Theory and application to reward shaping[C].\ Icml.\ 1999, 99: 278-287.

    \bibitem{ref28} Eysenbach B, Gupta A, Ibarz J, et al.\ Diversity is all you need: Learning skills without a reward function[J].\ arXiv preprint arXiv:1802.06070, 2018.

    \bibitem{ref29} Campbell M, Hoane Jr A J, Hsu F. Deep blue[J].\ Artificial intelligence, 2002, 134(1-2): 57-83.

    \bibitem{ref30} Sadeghi F, Levine S. Cad2rl: Real single-image flight without a single real image[J].\ arXiv preprint arXiv:1611.04201, 2016.
\end{thebibliography}
